\documentclass[abstract,final]{nldl}

\usepackage{mathtools}
\usepackage{graphicx}
\usepackage{booktabs}
\usepackage{enumitem}
\usepackage{algorithm}
\usepackage{float}
\usepackage{algorithmic}
\usepackage{amssymb}

\usepackage{listings}
\usepackage{hyperref}
\usepackage{url}
\hypersetup{colorlinks,
            linkcolor = BrickRed,
            citecolor = NavyBlue,
            urlcolor  = Magenta!80!black}

\title{Fragment-Level Graph Encoding for Interpretable Molecular
Property Prediction}
\author[1]{Rafael Lopes Almeida}
\author[1]{Ina P\"ohner\thanks{Corresponding Author.}}
\affil[1]{School of Pharmacy, University of Eastern Finland, 70211 Kuopio, Finland}
\affil[ ]{\texttt{\{rafael.lopes, ina.pohner\}@uef.fi}}

\begin{document}
\maketitle

\begin{abstract}
Graph neural networks (GNNs) are among the standard tools for molecular
property prediction, but the atom-level explanations of their
predictions are hard for medicinal chemists to act on. We propose
a change of input representation rather than a new model: molecules
are encoded as fragment-level graphs from BRICS decomposition,
where each node is a synthetically meaningful fragment whose
features are the element-wise sum of its atoms' features. Because
only the representation changes, any message-passing GNN and any
post-hoc attribution method produces explanations at the resolution
of functional groups instead of single atoms. Across four architectures,
four prediction tasks, and two explainers under scaffold splits,
fragment-level encoding matches atom-level test accuracy on roughly
five-fold smaller graphs. Integrated-Gradients attributions
recover known pharmacophores above chance on two enzyme targets,
namely BACE at 1.5 times the chance rate, and PTR1 at 3.2 times.
A like-for-like comparison against an aggregated atom-level
baseline shows that post-hoc aggregation is not a stable substitute:
its BACE hit rate range both exceeds and falls below
the fragment model's hit rate depending on two arbitrary aggregation choices,
whereas fragment-level training yields a single convention-free
attribution at no predictive cost. The code is available at: \url{https://github.com/4D-Lab/frame}

\end{abstract}

\section{Introduction}

Molecular property prediction in drug discovery supports computational
prioritization of compounds before committing costly synthesis and assay
resources. Graph neural networks (GNNs) have become a popular
choice because they operate directly on the molecular graph and
learn structural features tied to biological activity
\citep{gilmer_neural_2017, yang_analyzing_2019}, reaching
state-of-the-art performance on MoleculeNet benchmarks
\citep{wu_moleculenet_2018}.

However, these models are still largely uninterpretable. When a model flags
a molecule as active, a medicinal chemist wants to understand why, and which
parts of the molecule drive it, not only whether the prediction is right. 
Post-hoc methods such as Integrated Gradients (IG)
\citep{sundararajan_axiomatic_2017}, GNNExplainer
\citep{ying_gnnexplainer_2019}, and attention-based attribution
\citep{velickovic_graph_2018} assign importance to individual
atoms and bonds. The difficulty is one of resolution: chemists
reason in functional groups, pharmacophores, and ring systems
\citep{erlanson_twenty_2016}, so an atom heatmap that ranks carbon 14
above carbon 12 gives little to act on during lead optimization.
Fragment-based drug design already assembles and optimizes molecules
at the fragment/building block level \citep{murray_rise_2009, erlanson_twenty_2016}.
Explanations at that same resolution would thus speak the language chemists
already use.

We close this gap with a molecular graph encoding built from
BRICS decomposition \citep{degen_on_2008}. BRICS (Breaking of
Retrosynthetically Interesting Chemical Substructures) applies sixteen
retrosynthetically motivated rules that cleave bonds at known
synthetic disconnections, so the resulting fragments are both
chemically meaningful and synthetically accessible. In our encoding,
each BRICS fragment is a single node whose feature vector is the
element-wise sum of its atoms' features, and inter-fragment edges
mark the cleavage sites. The graph shrinks from dozens of atom nodes
to a handful of fragment nodes while keeping the original feature
information, and the granularity of any attribution method rises
from atoms to interpretable substructures.

The contribution is a change in data representation, not a new
architecture or a new explainer. Swapping atom-level graphs for
fragment-level graphs coarsens the graph and re-expresses GNN
explanations over chemically meaningful units, with
no architectural modifications, custom explainers, or extra training
objectives, and it drops into an existing pipeline as a preprocessing
step. We evaluate four architectures to demonstrate that the behavior
holds across different message-passing strategies, and two mechanistically
different explainers to probe whether it is an artifact of one
attribution method. Our contributions are:

\begin{enumerate}[leftmargin=1.4em, itemsep=1pt, topsep=2pt]
  \item A BRICS fragment-level graph encoding that lifts GNN explanation
  resolution from atoms to synthesizable substructures, with no change
  to the model or the explainer, usable as a drop-in preprocessing step.

  \item A quantitative evaluation protocol that pairs predictive metrics
  with three explanation metrics (fragment hit rate against known
  pharmacophores, attribution concentration, and cross-explainer
  agreement), all under scaffold splits, including an aggregated
  atom-level control that isolates the representation change from
  post-hoc aggregation.

  \item Evidence across four prediction tasks, four architectures, and two
  explainers that fragment encoding preserves predictive accuracy on
  far smaller graphs and recovers known pharmacophores above chance,
  while delineating where it helps (localized drug target interactions)
  and where it does not (a globally driven property).
\end{enumerate}

\section{Background \& Related Work}
\textbf{GNNs for molecular property prediction.}
GNN-based prediction methods have grown quickly since message passing was
introduced for quantum chemistry \citep{gilmer_neural_2017}. The
main families differ in how they aggregate neighbor information:
Graph attention (GAT) learns attention over neighbors \citep{velickovic_graph_2018},
with GATv2 removing a static-attention limitation \citep{brody_how_2022}.
Graph isomorphism networks (GINs) maximize discriminative power in the 
Weisfeiler--Leman hierarchy through sum aggregation \citep{xu_how_2019}.
Principal Neighborhood Aggregation (PNA) combines
several aggregators with degree-based scalers to overcome the limits
of any single aggregator \citep{corso_principal_2020}, and AttentiveFP
adds graph-level attention pooling tailored to molecules
\citep{xiong_pushing_2020}. All assume atoms are nodes and bonds are edges.

\textbf{Molecular representations and fragment-level methods.}
The standard molecular graph places each atom as a node and each bond
as an edge \citep{yang_analyzing_2019, wu_moleculenet_2018}. Alternatives
include fingerprints \citep{rogers_extended-connectivity_2010},
SMILES-based models \citep{weininger_smiles_1988}, and 3-D encodings
\citep{schutt_schnet_2018}. Fragment-level representations have drawn
interest for both prediction and generation: FragNet uses separate
message-passing channels for atoms, bonds, fragments, and inter-fragment
links \citep{panapitiya_fragnet_2026}. Fragment biases improve
expressivity and generalization \citep{wollschlager_expressivity_2024},
and subgraph-conditioned information-bottleneck pretraining learns
fragment-aware representations \citep{hoang_pre-training_2025}. These
share the intuition that fragments carry useful signal but also share a
requirement for custom architecture or modified message passing. Our
approach stays at the representation level and feeds unmodified
architectures, so it requires no change to an existing training pipeline.

\textbf{Interpretability: the two closest works.}
Substructure mask explanation (SME) segments molecules with BRICS,
Murcko, and functional-group rules, then explains an atom-level GNN by
masking each substructure and measuring the change in prediction
\citep{zhang_chemistry-intuitive_2023}. It is a post-hoc technique on a
model trained with standard atom-level graphs. SEAL instead modifies the
GNN to reduce inter-fragment message passing, yielding more causally
faithful fragment attributions \citep{musial_fragment-wise_2025}.

Our method differs from both in where it intervenes: we change the input
representation rather than applying post-hoc masking (unlike SME) or
altering the architecture (unlike SEAL), so it works with any architectures
and any attribution method. The aggregated atom-level baseline in
Section~\ref{sec:explain} is an SME-style control that tests whether the
representation change adds anything compared to post-hoc aggregation.

\textbf{Explainability methods and evaluation.}
Post-hoc GNN explainers fall into gradient-, perturbation-, and
decomposition-based families
\citep{sundararajan_axiomatic_2017, selvaraju_grad-cam_2017,
smilkov_smoothgrad_2017, ying_gnnexplainer_2019,
yuan_explainability_2023}. Applied to molecular GNNs, IG can highlight
features consistent with known pharmacophores \citep{jimenezluna_coloring_2021},
and quantitative studies find that gradient methods tend to give the
most reliable atom-level attributions \citep{rao_quantitative_2022,
wellawatte_model_2022}. Because our contribution is orthogonal to
the choice of explainer, we run two mechanistically different methods:
Integrated Gradients (IG, gradient-based) and GNNExplainer (perturbation-based).

\textbf{Scaffold splits.}
Random splits let structurally similar molecules land on both sides
of the train/test boundary and inflate generalization estimates
\citep{yang_analyzing_2019, chithrananda_chemberta_2020}. Murcko
scaffold splits \citep{bemis_properties_1996} keep molecules sharing
a core ring system together, forcing generalization across scaffold
boundaries. They remain generous relative to truly dissimilar test
sets \citep{tossou_scaffold_2024} but are the standard proxy for
novelty-based evaluation, and all experiments herein use them.

\section{Method}
\subsection{Fragment-level graph construction}

\textbf{Atom-level baseline.}
Each molecule is a graph with atoms as nodes and bonds as edges.
Node features are a 34-dimensional vector: element one-hot (10),
degree (6), formal charge (1), radical electrons (1), hybridization
(7), aromaticity (1), total hydrogen count (5), chirality flag (1),
and CIP stereochemistry (2). Edge features are 10-dimensional:
bond type (4), conjugation (1), ring membership (1), and stereochemistry
(4). Graphs are built from SMILES with RDKit \citep{landrum_rdkit_2016}.

\textbf{Fragment-level graph.}
Each molecule is decomposed with BRICS as implemented in RDKit,
which makes the decomposition deterministic. Given atom feature
vectors \(x_i \in \mathbb{R}^{34}\), the fragment node feature is
the element-wise sum over the atoms assigned to fragment \(f\):

\begin{equation}
  h_f = \sum_{i \in f} x_i,
  \label{eq:sum}
\end{equation}

The procedure is: (1) find BRICS-cleavable bonds, (2) break them without
adding dummy atoms, (3) map each atom to its resulting fragment, (4) sum
atom features per fragment as in Eq.~\eqref{eq:sum}, (5) add an
undirected edge between two fragments for each cleaved bond, and (6) give
every fragment--fragment edge the same constant scalar attribute, so that
no bond-type information enters the fragment graph.

\begin{figure}[tbh]
  \centering
  \includegraphics[width=\linewidth]{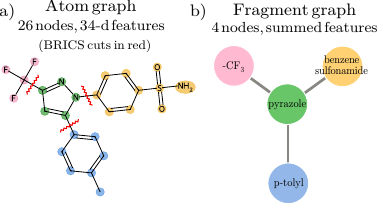}
    \caption{Atom-level versus BRICS fragment-level graph construction for
    celecoxib. (a) The atom graph: every atom is a node (26 nodes,
    34-dimensional features) and every bond an edge, with nodes ringed and
    coloured by BRICS fragment and cut bonds in red. (b) The fragment-level
    graph (4 fragment nodes with element-wise summed features), each node a
    recognizable substructure.}
  \label{fig:construction}
\end{figure}

Sum aggregation, rather than
mean or max, keeps sensitivity to fragment size, a real chemical
distinction. Inter-fragment edges are uniform because the chemistry of
the connection is already captured by the BRICS rules. Molecules with no
BRICS-cleavable bonds cannot be decomposed and are excluded, and the
exclusion rate is reported per dataset (Section~\ref{sec:setup}).
Figure~\ref{fig:construction} illustrates the construction: a drug-like
molecule with 20--40 heavy atoms becomes a graph of roughly 4--8 nodes,
so there are fewer message-passing operations and fewer attribution
targets, each a recognizable substructure.

\subsection{Data splitting}
All datasets use a Murcko scaffold split at 80/10/10. Molecules sharing
a Bemis--Murcko scaffold \citep{bemis_properties_1996} were always
assigned to the same split. Scaffold groups are sorted largest-first
and assigned in turn. The procedure is deterministic and prevents
scaffold leakage. Random splits were avoided because they inflate
generalization by placing similar molecules on both sides of the
boundary \citep{yang_analyzing_2019}.

\subsection{Architectures and training}
Four architectures were tested to evaluate whether the encoding works
regardless of the message passing approach, not to find a 'winner':
AttentiveFP with graph-level attention pooling \citep{xiong_pushing_2020},
GAT/GATv2 with attention and edge features \citep{velickovic_graph_2018,
brody_how_2022}, PNA combining multiple aggregators with degree scalers
\citep{corso_principal_2020}, and GIN with learnable epsilon, sum
aggregation, and batch normalization in each multilayer perceptron
\citep{xu_how_2019}.
Graph-level predictions come from global pooling of the final node
embeddings followed by a single output unit. The training pipeline is
identical for both encodings, so any difference is attributable to
the representation alone: Lookahead \citep{zhang_lookahead_2019}
over Adam \citep{kingma_adam_2015} with a cosine schedule and
warmup \citep{loshchilov_sgdr_2017}, class-weighted loss for
classification and mean squared error (MSE)/Huber for regression,
edge-dropout and feature-masking augmentation, and early stopping on
the validation metric. Reported test metrics are the mean $\pm$
standard deviation of the winning configuration retrained across
five seeds. Hyperparameters are tuned with Optuna's TPE sampler
\citep{akiba_optuna_2019} (150 trials per experiment).
Search spaces and best configurations are in
Appendices~\ref{app:search}--\ref{app:best}.

\subsection{Explainability and its evaluation}
\label{sec:explain}

\textbf{Attribution methods.}
We use two mechanistically different methods so findings do not hinge
on one technique. IG \citep{sundararajan_axiomatic_2017},
via Captum \citep{kokhlikyan_captum_2020}, integrates the output
gradient along a straight-line path from a zero baseline to the input.
GNNExplainer \citep{ying_gnnexplainer_2019}, via PyTorch Geometric,
learns a soft node/edge mask that maximizes mutual information with
the prediction. For both methods and encodings, per-node attribution
vectors are reduced to scalar scores by summing over feature dimensions.
In the fragment-level case each score maps to a recognizable substructure.

\textbf{Aggregated atom-level baseline.}
An obvious question is whether fragment-level graphs are needed at all:
one could train an atom-level GNN, compute atom attributions, and sum
them per fragment. This baseline is implemented directly with the same
architecture and training, so comparing it against the native
fragment-level model isolates whether the representation change,
rather than the aggregation step, drives any improvement. This is the
SME-style control of Section~2. Aggregation is not uniquely defined,
so we report all four combinations of two choices: summing signed
attributions (the default, under which opposing atom contributions
cancel each other) or their magnitudes, each either unnormalized
(rewarding large fragments) or divided by the fragment's atom count.
The native fragment model needs no such choice.

\textbf{Quantitative metrics.}
Beyond qualitative maps we compute three metrics over the test set.
The \emph{fragment hit rate} (plausibility) is the fraction of molecules
whose top-attributed fragment belongs to a pharmacophorically relevant
class defined per target from published structure--activity relationships (SAR).
Class membership is decided on the intact molecule: each SMARTS pattern is
matched against the parent structure, and a fragment is credited with a
class when all atoms of a match fall inside it. Matching the fragment
in isolation would give a wrong answer, because breaking a bond frees a
valence that RDKit refills with hydrogen, causing hydrogen-count-sensitive
patterns to match an artificial atom environment. Appendix~\ref{app:matching}
quantifies the effect. A class label therefore attests that a fragment
carries a recognition element drawn from published SAR, not that the
fragment is itself a published pharmacophore around which a chemical
series was designed. For BBBP (blood brain barrier penetration), which
has no localized pharmacophore, we use a directional criterion based on
topological polar surface area (TPSA; a BBB\textsuperscript{+} prediction
should attribute to a low-TPSA fragment, BBB\textsuperscript{-} to a
high-TPSA one), with the fragment's TPSA summed from the parent molecule's
per-atom contributions for the same reason.

\emph{Attribution concentration} (compactness) asks whether an
explanation points at anything in particular. Every attribution method
returns one number per unit, and those numbers can be sharp or flat.
When all of a molecule's fragments receive similar scores, the map says
only that everything matters a little, which gives a chemist nothing to
act on even if each score is individually correct. When one fragment
holds more of the attribution mass than the rest combined, the map names
a substructure to keep or to replace. We summarize this with the Gini
coefficient of the per-molecule importance vector, computed on absolute
values so that a strongly negative contribution counts as informative
rather than canceling a positive one. It is 0 when every unit receives
an equal share and 1 when one unit holds everything. The hit rate asks
whether the top-ranked fragment is chemically sensible while concentration
asks whether there is a dominant fragment that drives the prediction.

\begin{figure}[tbh]
  \centering
  \includegraphics[width=\linewidth]{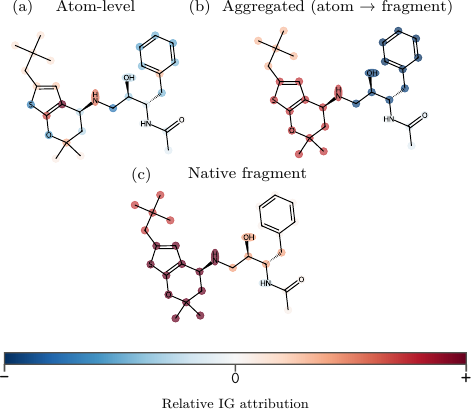}
  \caption{Integrated Gradients (IG) attribution at three
  resolutions for a correctly-predicted BACE inhibitor.
  (a) Raw atom-level attribution is fine-grained but noisy. (b) Atom
  attributions aggregated into fragments (SME-style control). (c) Native
  fragment-level attribution concentrates on the aminothienopyran
  recognition element. Each panel is scaled to its own maximum.}
  \label{fig:workflow}
\end{figure}

\emph{Cross-explainer agreement} is the per-molecule Spearman correlation
between IG and GNNExplainer fragment rankings. Fragment importance is
visualized as a color-coded heatmap on the 2-D structure, with every atom
taking the importance of its parent fragment (Figure~\ref{fig:workflow}).

\section{Experiments \& Results}

\subsection{Setup}
\label{sec:setup}

Our three prediction targets cover two distinct scenarios at once: they range from
localized to globally distributed SAR, and from public benchmarks of a few thousand
molecules to a few hundred data points manually compiled from the literature. 
One target, BACE, is used in both classification and regression settings,
resulting in four prediction tasks.

\emph{BACE} (MoleculeNet, $\beta$-secretase~1) has extensive published
SAR (hydroxyethylamine transition-state mimics engaging catalytic
Asp32/Asp228, hydrophobic S1/S2 groups, basic S3 amines
\citep{ghosh_developing_2012, stamford_inhibitors_2007}), making it
the sharpest test of pharmacophore recovery.

\emph{BBBP} \citep{wu_moleculenet_2018} labels blood--brain-barrier
permeability, a property governed by global physicochemistry
(lipophilicity, polar surface area) rather than a binding site
\citep{pardridge_drug_2012}, so it is the limiting case.

\emph{PTR1} (pteridine reductase~1 from \emph{Trypanosoma brucei},
regression) is an NADPH-dependent reductase responsible for the 
resistance to inhibitors of the established anti-infective target
dihydrofolate reductase, which makes it a co-target in human African
trypanosomiasis \citep{sienkiewicz_trypanosoma_2010,
ong_dissecting_2011}. Its active site sandwiches an inhibitor ring
between the nicotinamide of the cofactor and a phenylalanine of the
protein \citep{dawson_structure_2006}.
The recognition elements are a 2,4-diaminopteridine, -quinazoline or
-pyrimidine head hydrogen-bonding to the catalytic serine and
tyrosine, the stacked aromatic ring itself, and a peripheral acidic or
hydrophobic tail.

Unlike BACE and BBBP, PTR1 has no curated public benchmark. 250 compounds 
were manually compiled from sixteen primary medicinal-chemistry
resources published between 2009 and 2022 \citep{khalaf_structurebased_2014,
linciano_enhancement_2019, pohner_multitarget_2022, borsari_profiling_2016,
spinks_design_2011, santucci_repurposing_2021}.
Appendix~\ref{app:ptr1sources} lists all 16 sources with their
per-paper compound counts. We include PTR1 as a representative of the data 
situation in an early-stage program against a neglected-disease target: 
a few hundred literature measurements rather than a benchmark with little 
prospect of obtaining more. The evaluation is correspondingly small. 
Scaffold splitting over 119 Bemis--Murcko scaffolds, 82 of them singletons, 
sends every multi-member scaffold group to training and leaves 25 molecules 
each for validation and test. Because of this, model selection is unreliable, 
as can be expected in a data-sparse early-stage program. Further details 
are provided in Appendix~\ref{app:scarcity}.

BACE classification is roughly balanced (46\% active), BBBP is skewed
toward permeable compounds (76\%), and the regression targets span a
pIC$_{50}$ range of 2.5--10.5 (BACE) and a pK$_\text{i}$ range of
3.2--10.0 (PTR1).
Table~\ref{tab:setup} summarizes the datasets. Primary metrics are
the Matthews correlation coefficient (MCC) for classification and the
concordance correlation coefficient (CCC) for regression
\citep{chicco_advantages_2020, lin_concordance_1989}, and the full
metric set is in Appendix~\ref{app:fullmetrics}. Results are the best
validation checkpoint evaluated on the held-out scaffold test set.

\begin{table}[tbh]
  \centering
  \caption{Datasets used. $N$ is the molecule count and Scaf.
  the number of unique Bemis--Murcko scaffolds. The two regression
  targets carry different constants: BACE is pIC$_{50}$ and PTR1 is
  pK$_\text{i}$. Graph sizes are in Table~\ref{tab:scaffold}.}
  \label{tab:setup}
  \resizebox{\linewidth}{!}{\begin{tabular}{llrrr}
\toprule
Dataset & Task & $N$ & Scaf. & Label \\
\midrule
BACE (cls) & cls & 1512 & 649 & 690/1512 pos (46\%) \\
BACE (reg) & reg & 1512 & 649 & 2.54--10.52 ($\mu$=6.52) \\
BBBP & cls & 1965 & 1020 & 1495/1965 pos (76\%) \\
PTR1 & reg & 250 & 119 & 3.19--10.00 ($\mu$=5.61) \\
\bottomrule
\end{tabular}
}
\end{table}

\subsection{Predictive performance}
Tables~\ref{tab:predictive_cls} and \ref{tab:predictive_reg} report the
primary test metric (mean $\pm$ std over five training seeds) for all
four architectures under both encodings, for the classification and
regression targets, respectively. Fragment-level encoding is
competitive with atom-level encoding throughout.
On the two BACE datasets the best fragment model
matches or exceeds the best atom model (0.587 vs.\ 0.530 MCC on
classification; on regression the two are level, 0.861 vs.\ 0.858 CCC, a
gap far inside the seed-to-seed spread that we do not read as a
difference), and it leads on PTR1 (0.439 vs.\ 0.300 CCC), whereas the best atom model
leads on BBBP (0.291 vs.\ 0.186 MCC).

One note is that every PTR1 configuration has a negative test $R^2$ 
($-0.09$ to $-1.91$; Appendix~\ref{app:fullmetrics}), so no model
beats a constant predictor in squared error on 25 held-out molecules.
CCC stays positive because it rewards rank-and-scale agreement
rather than calibration. We therefore read the PTR1 column as a
comparison between the two encodings on a common split, not as
evidence that either predicts potency, and we return to the
reason in Appendix~\ref{app:scarcity}.

Aggregating over the sixteen dataset--architecture pairs, the fragment
encoding wins eleven and loses five, with a mean primary-metric change
of $+0.037$ and comparable seed-to-seed variance (mean test standard
deviation 0.040 for atom and 0.045 for fragment encodings). No architecture
dominates, and the atom/fragment ranking is inconsistent across architectures:
GIN improves under the fragment encoding on all four datasets, while
AttentiveFP is strongest at the atom level on BBBP but, on PTR1,
the weakest atom-level architecture (0.267, against 0.300 for PNA) and
the strongest fragment-level one (0.439). This is expected when the
representation, not the architecture, is the variable of interest.

These accuracies are obtained on graphs with roughly five times fewer
nodes (34.1 $\rightarrow$ 7.2 for BACE, Table~\ref{tab:scaffold})
at comparable training cost, cutting the number of attribution targets
per explanation from dozens of atoms to a handful of substructures.
BBBP is hard for every model (MCC $\le 0.30$), consistent with a
globally driven property.

\begin{table}[tbh]
  \centering
  \caption{Classification: primary test metric (MCC) per architecture and
  encoding, atom (a) vs.\ fragment (f). Values are
  mean\,$\pm$\,std over five training seeds, with the best mean per
  block in bold. Full metrics in Appendix~\ref{app:fullmetrics}.}
  \label{tab:predictive_cls}
  \resizebox{\linewidth}{!}{\begin{tabular}{lcccc}
\toprule
 & \multicolumn{2}{c}{BACE (cls)} & \multicolumn{2}{c}{BBBP} \\
Backbone & MCC$_\mathrm{a}$ & MCC$_\mathrm{f}$ & MCC$_\mathrm{a}$ & MCC$_\mathrm{f}$ \\
\midrule
AttentiveFP & 0.530{\scriptsize$\,\pm\,0.073$} & \textbf{0.565}{\scriptsize$\,\pm\,0.050$} & \textbf{0.291}{\scriptsize$\,\pm\,0.040$} & 0.186{\scriptsize$\,\pm\,0.042$} \\
GATv2 & 0.439{\scriptsize$\,\pm\,0.057$} & \textbf{0.552}{\scriptsize$\,\pm\,0.058$} & 0.140{\scriptsize$\,\pm\,0.020$} & \textbf{0.176}{\scriptsize$\,\pm\,0.065$} \\
GIN & 0.385{\scriptsize$\,\pm\,0.087$} & \textbf{0.476}{\scriptsize$\,\pm\,0.104$} & 0.163{\scriptsize$\,\pm\,0.050$} & \textbf{0.183}{\scriptsize$\,\pm\,0.051$} \\
PNA & 0.453{\scriptsize$\,\pm\,0.085$} & \textbf{0.587}{\scriptsize$\,\pm\,0.030$} & \textbf{0.175}{\scriptsize$\,\pm\,0.027$} & 0.151{\scriptsize$\,\pm\,0.057$} \\
\bottomrule
\end{tabular}}
\end{table}

\begin{table}[tbh]
  \centering
  \caption{Regression: primary test metric (CCC) per architecture and
  encoding, atom (a) vs.\ fragment (f). Values are
  mean\,$\pm$\,std over five training seeds, with the best mean per
  block in bold. Full metrics in Appendix~\ref{app:fullmetrics}.}
  \label{tab:predictive_reg}
  \resizebox{\linewidth}{!}{\begin{tabular}{lcccc}
\toprule
 & \multicolumn{2}{c}{BACE (reg)} & \multicolumn{2}{c}{PTR1} \\
Backbone & CCC$_\mathrm{a}$ & CCC$_\mathrm{f}$ & CCC$_\mathrm{a}$ & CCC$_\mathrm{f}$ \\
\midrule
AttentiveFP & 0.846{\scriptsize$\,\pm\,0.013$} & \textbf{0.861}{\scriptsize$\,\pm\,0.007$} & 0.267{\scriptsize$\,\pm\,0.027$} & \textbf{0.439}{\scriptsize$\,\pm\,0.033$} \\
GATv2 & \textbf{0.818}{\scriptsize$\,\pm\,0.009$} & 0.814{\scriptsize$\,\pm\,0.031$} & \textbf{0.261}{\scriptsize$\,\pm\,0.061$} & 0.229{\scriptsize$\,\pm\,0.034$} \\
GIN & 0.797{\scriptsize$\,\pm\,0.009$} & \textbf{0.800}{\scriptsize$\,\pm\,0.056$} & 0.298{\scriptsize$\,\pm\,0.035$} & \textbf{0.379}{\scriptsize$\,\pm\,0.053$} \\
PNA & \textbf{0.858}{\scriptsize$\,\pm\,0.007$} & 0.845{\scriptsize$\,\pm\,0.010$} & 0.300{\scriptsize$\,\pm\,0.044$} & \textbf{0.377}{\scriptsize$\,\pm\,0.040$} \\
\bottomrule
\end{tabular}
}
\end{table}

\subsection{Explanation quality}
\label{sec:expq}
Table~\ref{tab:explain} reports explanation quality for the three
approaches on a representative architecture per target, chosen as the
architecture with the highest mean primary metric across the two
encodings: AttentiveFP for BACE and PTR1, GIN for BBBP.

\begin{table}[tbh]
  \centering
  \caption{Explanation quality by approach. Hit: fragment hit rate
  against known pharmacophores (BBBP: directional TPSA), with the
  reachable ceiling imposed by BRICS given below the BACE and PTR1
  blocks. BBBP has no discrete pharmacophore and thus no ceiling.
  Hit$^{\ast}_{\text{IG}}$: the same IG hit rate normalized by that
  ceiling, i.e.\ the rate among molecules that can register a hit at all,
  which makes targets with different ceilings comparable; interval
  estimates for it are given in Section~\ref{sec:explain}.
  The best value per dataset block is set in bold, and exact ties are
  left unbolded.
  Gini: attribution concentration, corrected by the maximum attainable
  at each vector length, so atom- and fragment-level values are
  comparable. $\rho_{\text{IG,GX}}$: per-molecule IG vs.\ GNNExplainer
  Spearman agreement. The aggregated row uses the default signed-sum
  convention, and Appendix~\ref{app:conventions} gives its full range.}
  \label{tab:explain}
  \resizebox{\linewidth}{!}{\begin{tabular}{llccccccc}
\toprule
Data & Approach & $n$ & Hit$_{\text{IG}}\uparrow$ & Hit$^{\ast}_{\text{IG}}\uparrow$ & Hit$_{\text{GX}}\uparrow$ & Gini$_{\text{IG}}\uparrow$ & Gini$_{\text{GX}}\uparrow$ & $\rho_{\text{IG,GX}}\uparrow$ \\
\midrule
BACE & Atom-level & 148 & \textbf{0.405} & \textbf{0.465} & 0.345 & 0.478 & 0.122 & -0.029 \\
 & Aggregated & 148 & 0.399 & 0.457 & \textbf{0.514} & \textbf{0.532} & \textbf{0.403} & 0.053 \\
 & Native frag. & 148 & 0.277 & 0.318 & 0.480 & 0.518 & 0.282 & \textbf{0.330} \\
\addlinespace[1pt]
& \emph{reachable ceiling} & & \emph{0.872} & & & & & \\
\midrule
BBBP & Atom-level & 166 & \textbf{0.723} & --- & \textbf{0.741} & 0.258 & 0.075 & 0.155 \\
 & Aggregated & 166 & 0.711 & --- & 0.723 & 0.455 & \textbf{0.434} & \textbf{0.812} \\
 & Native frag. & 166 & 0.693 & --- & 0.711 & \textbf{0.484} & 0.249 & 0.270 \\
\midrule
PTR1 & Atom-level & 23 & 0.783 & 1.000 & 0.739 & 0.751 & 0.157 & 0.486 \\
 & Aggregated & 23 & 0.783 & 1.000 & 0.783 & \textbf{0.848} & \textbf{0.564} & 0.688 \\
 & Native frag. & 23 & 0.783 & 1.000 & 0.783 & 0.732 & 0.324 & \textbf{0.837} \\
\addlinespace[1pt]
& \emph{reachable ceiling} & & \emph{0.783} & & & & & \\
\bottomrule
\end{tabular}
}
\end{table}

\begin{figure}[tbh]
  \centering
  \includegraphics[width=\linewidth]{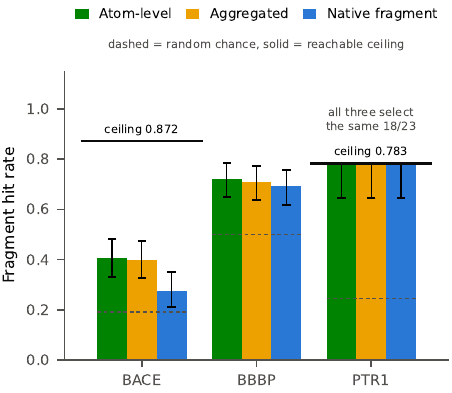}
  \caption{Fragment hit rate using IG by target and approach, with 95\%
  intervals conditional on reachability: for BACE and PTR1 the interval is
  estimated over the molecules that can register a hit and rescaled by the
  ceiling, so it cannot exceed it; BBBP has no ceiling and uses a Wilson
  interval on the raw proportion. Dashed lines mark the random-chance rate
  per target, solid rules the reachable ceiling. BACE and PTR1 attributions
  exceed chance for all approaches. The three PTR1 bars coincide at the
  ceiling, where all three approaches select the same 18 of 23 molecules
  (Appendix~\ref{app:scarcity}). BBBP uses the directional TPSA
  criterion.}
  \label{fig:hitrate}
\end{figure}

\textbf{Attributions recover pharmacophores above chance.}
On BACE, 19.1\% of fragments belong to a pharmacophoric class, and the
top-attributed fragment is pharmacophoric for 28--41\% of the test molecules
depending on the approach (Figure~\ref{fig:hitrate}): an enrichment of
1.5-fold for the native fragment model and 2.1-fold for atom-level
attribution. The native model reaches 0.277 and, with atom attributions
aggregated per fragment under the default convention, 0.399.
Fragment-level attributions therefore surface medicinally meaningful
substructures at dataset scale rather than only on curated examples,
though the margin over chance is narrower than the raw rates suggest.

A per-class breakdown shows what limits these numbers. A fragment can
only be credited with a motif that fits inside it, so the metric is
bounded by what the decomposition leaves whole. On the BACE test split,
the hydroxyethylamine transition-state mimic, the pharmacophore a
chemist would name first, is present in 52\% of the molecules and 
lands inside a single fragment in none of the 148, because it spans a
bond BRICS breaks during construction. The benzylic S2 class survives 
in 21\% of the cases, whereas the basic amine and the S1 alkyl class 
survive intact whenever they occur.

PTR1 is the counter-example, and it isolates the decomposition as the
cause. Across all 239 decomposable PTR1 molecules, the pterin head is
present in 34.7\% and survives BRICS intact in 88.0\% of those, the
stacked aromatic in 18.8\% and 68.9\%, and both tail classes in 4.2\%
and 100\%. On the 23-molecule test split every class that occurs
survives. BRICS cleaves BACE's defining motif by construction and
leaves PTR1's untouched, so the gap between the two targets is a
property of the decomposition approach rather than of the models.

The ceilings follow: at most 0.872 of BACE and 0.783 of PTR1 test
molecules can register a hit under any method. On PTR1, all three
approaches reach 0.783, which matches the ceiling exactly, i.e.\ a
ceiling-normalized rate of 1.000, against a chance rate of 0.245.
Because the hit rate cannot exceed the ceiling, a Wald interval on the raw
proportion is misspecified here: it is symmetric and unbounded above, and at
$n=23$ with the estimate on the boundary it returns [0.614, 0.951], whose upper
limit lies above the 0.783 ceiling. We therefore estimate the hit rate
\emph{conditional on reachability}. Of the 23 test molecules 18 are reachable
and all 18 are hit, so the conditional rate is 1.000 with a Wilson 95\% interval
of [0.824, 1.000], equivalently [0.645, 0.783] after rescaling by the ceiling.
Both exclude the 0.245 chance rate, resulting in a 3.2-fold enrichment. Above-chance recovery therefore holds on both
enzyme targets. However, all three approaches select the \emph{same} 18 of 23
molecules, so the metric saturates rather than separating them.
Appendix~\ref{app:scarcity} gives the details and shows that the claim
is robust to how the pharmacophore classes are drawn.

The encoding is agnostic to the decomposition, so the plausibility
metric inherits whatever boundaries the chosen scheme imposes.

\textbf{Post-hoc aggregation is not a stable substitute for
fragment-level training.}
Summing atom attributions into fragments is under-specified:
it requires two choices that no theory fixes, whether to sum signed
attributions or their magnitudes, and whether to normalize by fragment
size. Both choices move the result. IG attributions change sign within 
a fragment, so a signed sum lets positive and
negative atoms cancel before the fragment score is formed. We measure
how much survives that cancellation as the ratio between a fragment's
signed sum and the total attribution its atoms carry, ignoring sign.

A fragment whose atoms all point the same way scores 1, one whose atoms
cancel scores near 0. The average is 0.73 on BACE and drops to 0.43 for
six-atom BACE fragments, so on the larger fragments the signed
convention discards most of the atom-level signal. Mixed-sign fragments
are common rather than exceptional at 52\% of 956 fragments. On PTR1 the
effect is far weaker, at 0.90 and 47\% of 102 fragments: its molecules
carry 4.4 fragments against BACE's 7.2, and smaller fragments hold
fewer atoms to cancel each other out. The instability therefore
scales with fragment size, and BACE exposes this, whereas PTR1 does not.
Summing magnitudes avoids the cancellation and raises the baseline
(0.399 to 0.426 on BACE), while normalizing by size removes the reward
for large fragments and cuts it to 0.223. Across the four combinations
the baseline covers 0.223--0.426 on BACE and 0.696--0.783 on PTR1
(Table~\ref{tab:conventions}). The native fragment model scores 0.277
on BACE, inside that range, and 0.783 on PTR1, where it ties the top of
the range because three of the five values coincide exactly.

The fragment-level model requires none of these choices: because it
predicts at fragment resolution, its attribution is the fragment
score, and it reports a single convention-free number. That is the
advantage of moving the representation rather than post-processing an
atom model, since the explanation then does not depend on a free
parameter the reader cannot audit. On BACE, aggregation exceeds native
attribution under two of the four conventions and falls below it under
the other two, but on PTR1 it never exceeds it and two conventions
coincide with it exactly. We draw no directional conclusion from this
comparison.

\textbf{Correcting for granularity changes the concentration ordering,
but not uniformly.}
The raw Gini coefficient favors whichever representation has more units,
since it cannot exceed $(n-1)/n$. Correcting for that bound 
(Section~\ref{sec:explain})
changes the ordering, though the direction is target-dependent. Native
attribution is the most concentrated on BBBP (0.484 against 0.455 for
aggregated and 0.258 for atom-level), and on BACE it exceeds the
atom-level baseline (0.518 vs.\ 0.478) while sitting just below
aggregation. On PTR1 it trails both (0.732 against 0.848 and 0.751).
Uncorrected, the same comparison ranks native last on BACE purely
because its vectors are five times shorter, so the correction is
necessary even where it does not favor the fragment model.
Concentration is what makes a map readable: placing attribution mass on
a handful of substructures rather than spreading it over dozens of
atoms is the point of the encoding. PTR1 reverses the ordering, which
is consistent with its granularity: 4.4 fragments leave a corrected
Gini little room to separate the three approaches.

\textbf{Explainer agreement is strongly target-dependent.}
GNNExplainer matches or exceeds IG on hit rate in all six fragment-level
cells (0.480 vs.\ 0.277 for native attribution on BACE, and an exact
tie at 0.783 on PTR1). How closely their rankings agree, however,
varies more than any other quantity we measure. Per-molecule Spearman
agreement for the native model is 0.33 on BACE, 0.27 on BBBP and 0.84
on PTR1, and across the nine dataset-approach cells it spans $-0.03$ to
0.84 with a median of 0.33. Agreement rises as the fragment count
falls: with 4.4 units per molecule there is little room for two
rankings to diverge, whereas BACE's 7.2 and BBBP's 4.5 units allow
genuine disagreement. On the benchmark-scale targets, the two methods
locate overlapping but distinct fragments, which is the reason to
report both rather than treat either as ground truth. Where they part
company most is concentration: GNNExplainer's corrected Gini falls
below IG's in every cell (0.282 vs.\ 0.518 for native attribution on
BACE), because its masks are shaped by a size-and-entropy penalty
rather than by the model's gradient alone.

We report IG in the main comparisons: it is deterministic and carries no
optimization hyperparameters of its own, and prior work finds
gradient-based attribution the more reliable family for molecular GNNs
\citep{rao_quantitative_2022, wellawatte_model_2022}. One ordering does
invert under GNNExplainer, where both fragment-level approaches outrank
atom-level attribution on BACE (0.514 and 0.480 against 0.345) rather
than trailing it. We do not lean on that reversal, since the diffuse
masks behind it are the same ones the Gini comparison penalizes, but it
does mean that the atom-level baseline's advantage on hit rate is specific 
to IG.

\subsection{Where it does not help: BBBP}
BBBP tests whether fragment attributions can decompose a global property
into local contributions. The directional TPSA hit rate is modestly
above chance (0.69 vs.\ 0.50), so the models do attribute to
physicochemically consistent fragments, but attribution is markedly less
concentrated than for the enzyme targets (corrected native Gini 0.48,
against 0.52 on BACE and 0.73 on PTR1): no small set of fragments
dominates the explanation, as is expected for a whole-molecule property.
BBBP also has the highest BRICS exclusion rate (12.9\% of molecules
have no cleavable bond, versus 0.3\% for BACE and 4.4\% for PTR1,
Appendix~\ref{app:scaffold}), because permeant molecules are often
small and rigid, a coverage gap that compounds the limitation.
Fragment-level explanations are thus most useful for targets with
localized SAR.

\section{Discussion}
\textbf{When fragment-level encoding works best.}
The encoding is most useful when activity is driven by localized
pharmacophoric interactions that map onto BRICS fragments. BACE and
PTR1 fit this pattern: both are enzymes with their binding sites made up
of distinct subpockets, and BRICS tends to cleave at the boundaries
between the recognition elements. Only BACE bears this out quantitatively
at 1.5-fold enrichment over 148 molecules, while PTR1 reaches 3.2-fold over
23, but the latter estimate rests on a small test set and cannot separate
the three approaches from one another. The two targets bracket the behavior:
BRICS destroys BACE's defining motif in every test molecule and preserves
PTR1's in 88\%. How high the plausibility metric can reach is therefore
set by how much of the pharmacophore survives fragmentation, not by the
model. Attribution on the resulting fragment graph then highlights
which substructures matter, in units a chemist can act on. Because
BRICS fragments are retrosynthetically valid, a high-importance fragment
is a real building block that can be looked up or set as a synthesis
target, unlike arbitrary graph-coarsening pieces. Appendix~\ref{app:maps}
shows further attribution maps for BACE and PTR1.

\textbf{What data scarcity does to the evaluation.}
Scarcity degrades the evaluation itself, separately from the
representation question, and Appendix~\ref{app:scarcity} separates the
two: PTR1 model selection and calibration both fail at this sample
size, while the attributions stay pharmacophorically sensible. Fragment
encoding is therefore defensible where data are scarce, but
fragment-level explanation comparisons at that scale are not.

\textbf{Limitations.}
The comparison against the aggregated baseline is convention-dependent
in both directions: we show the baseline is unstable, but we do not
claim a pharmacophore-recovery advantage for native attribution either,
and establishing a fidelity advantage would require a causal faithfulness
study similar to the SEAL targets \citep{musial_fragment-wise_2025}.

A stronger limitation comes from re-scoring attributions at atom
resolution against pharmacophore matches taken on the intact molecule,
which drops the fragment boundaries from the evaluation entirely. On
BACE, no approach separates pharmacophore atoms reliably: across the
twelve class--approach cells the AUC spans 0.26--0.61 with a median of
0.47, and 67\% of values lie within 0.1 of chance, for native and
aggregated attribution alike. PTR1 gives a mixed picture. Its two
well-populated classes do separate, the pterin head at 0.88--0.94 and
the stacked aromatic at 0.38--0.65, but its two tail classes occur in a
single test molecule each, so their AUCs are meaningless. Across both
targets, atom-resolution separation tracks how much of the test set a
class covers, which means the plausibility metric itself, and not only
the representation, bounds what these experiments can settle.

Three further limits hold generally. Molecules with no BRICS-cleavable
bonds are excluded, a coverage gap for small or rigid structures. The
choice of BRICS is a modeling decision, and our survival measurements
show that it splits the defining BACE motif in every test molecule, so a
coarser decomposition, such as RECAP \citep{lewell_recap_1998},
or a scheme that protects matched groups from cleavage, is the natural
next step and would raise the measurable ceiling for every method. Summing
atom features within a fragment discards intra-fragment detail, so two
fragments with the same composition but different connectivity collapse
to the same vector. Finally, how closely the two explainers agree on
which fragments matters and varies by target rather than holding at 
one level (median $\rho = 0.33$, spanning $-0.03$ to 0.84), so ``any
explainer'' holds for the aggregate metrics but not for the fragment
ranking of an individual molecule, which should be read as
explainer-dependent.

\textbf{Practical implications.}
The fragment map reads directly as a lead-optimization worklist.
For the inhibitor in Figure~\ref{fig:workflow}, native attribution
ranks the aminothienopyran recognition element highest and the acetamido-
and benzyl-periphery lowest, so a chemist would preserve the top
fragment and treat the low-attribution caps as replacement candidates.
Because each cap is a BRICS building block, the replacement is a real,
synthetically accessible move rather than an abstract atom-level
cue. Beyond single molecules, fragment attributions can be averaged 
across a compound series to build fragment-level SAR tables, which adds a
quantitative signal to traditional SAR and connects prediction and
explanation to synthesis planning. We omit graph transformers by design.
The benefit is expected to extend to them, since they consume the same
node-level inputs, but we leave that to future work.

\section{Conclusion}
We described a fragment-level molecular graph encoding based on BRICS
decomposition. Making each retrosynthetically valid fragment a single
node lets any attribution method produce explanations at the resolution
of synthesis building blocks rather than atoms, with no change to the 
architecture or the explainer. Across four prediction tasks, four architectures, 
and two explainers under scaffold splits, fragment encoding matches atom-level
test accuracy on roughly five-fold smaller graphs, and IG attributions recover 
known pharmacophores above chance on both enzyme targets, at 1.5 times the 
chance rate on BACE and 3.2 times on PTR1.
A like-for-like comparison with an aggregated atom-level baseline shows
that post-hoc aggregation is not a stable substitute, since its hit rate on BACE
moves from below to above the fragment model's under two
arbitrary aggregation choices, so the benefit of moving the
representation is a convention-free explanation at competitive
accuracy. We do not claim dominance: across the four targets fragment
encoding trades wins with atom-level encoding on prediction, and on
plausibility the three approaches behave alike once the reachable ceiling
is accounted for. The contribution is an additional, convention-free
representation strategy for explainable molecular GNNs, not a uniformly
superior one. Future work includes alternative fragmentation schemes that
preserve functional groups, learned fragment aggregation via subgraph
GNNs, extension to graph transformers, and a causal faithfulness evaluation.

\section*{Acknowledgments}
The authors wish to acknowledge CSC – IT Center for Science, Finland, for
computational resources.

\section*{LLM Disclosure}
In this work, we used Claude Opus 4.7 through the UEF AI Assistant to check
spelling and grammar and to rephrase or rewrite sentences so they read more
clearly. We did not use LLMs for generating research ideas, analyzing
data, or producing results, and other disclosure categories are not applicable to
this work. We reviewed all AI-assisted edits, compared each suggested
change against our original draft, and only accepted changes that preserved 
our intended meaning. We take full responsibility for the final content of this
work, including any text produced or edited with the help of LLMs.

\printbibliography

\appendix
\onecolumn
\section{Hyperparameter search spaces}
\label{app:search}
All experiments share the Optuna/TPE search space in
Table~\ref{tab:search} (150 trials each, batch size 32, three tuning
seeds), with one exception: PTR1 uses batch size 16 and five tuning
seeds, because 200 training molecules give only thirteen optimizer
steps per epoch and its validation split is so small that a
single seed is unreliable. Every winning PTR1 configuration falls
inside the ranges tabulated here. The GAT variant (static/dynamic
attention) is itself a tuned choice, so some winning GAT
configurations use GATv1.
\begin{table}[h]
  \centering
  \caption{Hyperparameter search space.}
  \label{tab:search}
  \begin{tabular}{ll}
\toprule
Hyperparameter & Range / choices \\
\midrule
hidden channels & [32, 256] \\
layers & [2, 4] \\
dropout & [0.2, 0.6] \\
pooling & \{mean, add, max\} \\
learning rate & [0.0001, 0.005] (log) \\
weight decay & [1e-06, 0.001] (log) \\
timesteps (AttentiveFP) & [1, 3] \\
heads (GAT) & [1, 4] \\
GATv2 (GAT) & \{True, False\} \\
towers (PNA) & \{1, 2, 4, 6, 8\} \\
\bottomrule
\end{tabular}
\end{table}
\newpage

\section{Best hyperparameter configurations}
\label{app:best}
Table~\ref{tab:best} lists the winning configuration per experiment
(h: hidden channels, L: layers, d: dropout, lr: learning rate, pooling,
plus architecture-specific settings).
\begin{table}[h]
  \centering
  \caption{Best configurations (atom / fragment).}
  \label{tab:best}
  \resizebox{0.86\linewidth}{!}{\begin{tabular}{lll}
\toprule
Experiment & Enc. & Best configuration \\
\midrule
BACE (cls)/AttentiveFP & A & h234, L4, d0.23, lr2.8e-03, timesteps2 \\
BACE (cls)/AttentiveFP & F & h95, L3, d0.26, lr3.1e-03, timesteps3 \\
BACE (cls)/GAT & A & h213, L3, d0.24, lr3.7e-03, add, heads2, v1 \\
BACE (cls)/GAT & F & h149, L4, d0.28, lr4.9e-03, mean, heads4, v1 \\
BACE (cls)/GIN & A & h126, L2, d0.55, lr2.3e-03, mean \\
BACE (cls)/GIN & F & h170, L3, d0.48, lr2.7e-03, max \\
BACE (cls)/PNA & A & h156, L3, d0.22, lr6.8e-04, add, towers4 \\
BACE (cls)/PNA & F & h219, L4, d0.35, lr1.6e-04, max, towers4 \\
\midrule
BACE (reg)/AttentiveFP & A & h198, L4, d0.23, lr2.3e-03, timesteps3 \\
BACE (reg)/AttentiveFP & F & h143, L3, d0.21, lr1.7e-03, timesteps3 \\
BACE (reg)/GAT & A & h206, L3, d0.23, lr2.6e-03, add, heads4, v2 \\
BACE (reg)/GAT & F & h256, L2, d0.20, lr2.7e-03, add, heads4, v2 \\
BACE (reg)/GIN & A & h112, L3, d0.60, lr3.5e-03, add \\
BACE (reg)/GIN & F & h201, L3, d0.24, lr2.1e-03, add \\
BACE (reg)/PNA & A & h208, L3, d0.23, lr8.6e-04, max, towers4 \\
BACE (reg)/PNA & F & h232, L2, d0.23, lr7.0e-04, add, towers6 \\
\midrule
BBBP/AttentiveFP & A & h136, L3, d0.30, lr1.9e-03, timesteps3 \\
BBBP/AttentiveFP & F & h176, L4, d0.21, lr2.7e-03, timesteps1 \\
BBBP/GAT & A & h166, L2, d0.22, lr3.9e-03, add, heads3, v1 \\
BBBP/GAT & F & h198, L3, d0.22, lr4.2e-03, add, heads2, v2 \\
BBBP/GIN & A & h90, L2, d0.59, lr1.1e-03, add \\
BBBP/GIN & F & h80, L4, d0.47, lr4.8e-03, mean \\
BBBP/PNA & A & h37, L3, d0.30, lr3.6e-03, add, towers8 \\
BBBP/PNA & F & h228, L3, d0.35, lr1.6e-03, mean, towers8 \\
\midrule
PTR1/AttentiveFP & A & h214, L4, d0.53, lr4.7e-03, timesteps1 \\
PTR1/AttentiveFP & F & h221, L4, d0.48, lr4.3e-03, timesteps1 \\
PTR1/GAT & A & h172, L3, d0.20, lr3.3e-03, mean, heads2, v1 \\
PTR1/GAT & F & h224, L4, d0.23, lr4.6e-03, add, heads1, v1 \\
PTR1/GIN & A & h186, L4, d0.40, lr8.6e-04, add \\
PTR1/GIN & F & h206, L3, d0.23, lr1.9e-03, mean \\
PTR1/PNA & A & h244, L3, d0.49, lr7.1e-04, max, towers1 \\
PTR1/PNA & F & h182, L4, d0.54, lr5.5e-04, max, towers1 \\
\bottomrule
\end{tabular}
}
\end{table}
\newpage

\section{Complete metric tables}
\label{app:fullmetrics}
Tables~\ref{tab:cls} and \ref{tab:reg} give all test metrics for every
architecture/encoding cell.

\begin{table}[h]
  \centering
  \caption{Classification test metrics (A: atom, F: fragment). Abbreviations:
  acc: accuracy; acc\_bal: balanced accuracy (the mean per-class recall); f1: 
  $F_1$ score; prec: precision; rec: recall (f1, prec and recall for the positive
  class); mcc: Matthews correlation coefficient \citep{chicco_advantages_2020};
  avg\_prec: average precision (area under the precision--recall curve); roc\_auc:
  area under the ROC curve. All except avg\_prec and roc\_auc are computed at a 
  fixed 0.5 probability threshold, while avg\_prec and roc\_auc use continuous
scores and are threshold-free.}
  \label{tab:cls}
  \resizebox{\linewidth}{!}{\begin{tabular}{llcccccccc}
\toprule
Dataset & Model / Enc. & acc & acc\_bal & f1 & prec & rec & mcc & avg\_prec & roc\_auc \\
\midrule
BACE (cls) & AttentiveFP/A & 0.777{\scriptsize$\,\pm\,0.074$} & 0.781{\scriptsize$\,\pm\,0.033$} & 0.666{\scriptsize$\,\pm\,0.050$} & 0.590{\scriptsize$\,\pm\,0.092$} & 0.790{\scriptsize$\,\pm\,0.059$} & 0.530{\scriptsize$\,\pm\,0.073$} & 0.747{\scriptsize$\,\pm\,0.025$} & 0.867{\scriptsize$\,\pm\,0.015$} \\
 & AttentiveFP/F & 0.815{\scriptsize$\,\pm\,0.032$} & 0.794{\scriptsize$\,\pm\,0.019$} & 0.688{\scriptsize$\,\pm\,0.032$} & 0.642{\scriptsize$\,\pm\,0.066$} & 0.750{\scriptsize$\,\pm\,0.039$} & 0.565{\scriptsize$\,\pm\,0.050$} & 0.760{\scriptsize$\,\pm\,0.027$} & 0.864{\scriptsize$\,\pm\,0.029$} \\
 & GAT/A & 0.726{\scriptsize$\,\pm\,0.067$} & 0.737{\scriptsize$\,\pm\,0.029$} & 0.605{\scriptsize$\,\pm\,0.037$} & 0.515{\scriptsize$\,\pm\,0.071$} & 0.761{\scriptsize$\,\pm\,0.090$} & 0.439{\scriptsize$\,\pm\,0.057$} & 0.651{\scriptsize$\,\pm\,0.055$} & 0.817{\scriptsize$\,\pm\,0.017$} \\
 & GAT/F & 0.816{\scriptsize$\,\pm\,0.025$} & 0.784{\scriptsize$\,\pm\,0.030$} & 0.678{\scriptsize$\,\pm\,0.041$} & 0.647{\scriptsize$\,\pm\,0.050$} & 0.715{\scriptsize$\,\pm\,0.056$} & 0.552{\scriptsize$\,\pm\,0.058$} & 0.745{\scriptsize$\,\pm\,0.014$} & 0.854{\scriptsize$\,\pm\,0.013$} \\
 & GIN/A & 0.662{\scriptsize$\,\pm\,0.086$} & 0.712{\scriptsize$\,\pm\,0.049$} & 0.575{\scriptsize$\,\pm\,0.054$} & 0.450{\scriptsize$\,\pm\,0.073$} & 0.820{\scriptsize$\,\pm\,0.063$} & 0.385{\scriptsize$\,\pm\,0.087$} & 0.615{\scriptsize$\,\pm\,0.064$} & 0.816{\scriptsize$\,\pm\,0.025$} \\
 & GIN/F & 0.737{\scriptsize$\,\pm\,0.097$} & 0.756{\scriptsize$\,\pm\,0.053$} & 0.633{\scriptsize$\,\pm\,0.068$} & 0.536{\scriptsize$\,\pm\,0.102$} & 0.800{\scriptsize$\,\pm\,0.050$} & 0.476{\scriptsize$\,\pm\,0.104$} & 0.632{\scriptsize$\,\pm\,0.145$} & 0.805{\scriptsize$\,\pm\,0.057$} \\
 & PNA/A & 0.717{\scriptsize$\,\pm\,0.090$} & 0.746{\scriptsize$\,\pm\,0.046$} & 0.617{\scriptsize$\,\pm\,0.056$} & 0.508{\scriptsize$\,\pm\,0.086$} & 0.810{\scriptsize$\,\pm\,0.060$} & 0.453{\scriptsize$\,\pm\,0.085$} & 0.720{\scriptsize$\,\pm\,0.005$} & 0.853{\scriptsize$\,\pm\,0.011$} \\
 & PNA/F & 0.830{\scriptsize$\,\pm\,0.013$} & 0.803{\scriptsize$\,\pm\,0.017$} & 0.703{\scriptsize$\,\pm\,0.022$} & 0.666{\scriptsize$\,\pm\,0.025$} & 0.745{\scriptsize$\,\pm\,0.037$} & 0.587{\scriptsize$\,\pm\,0.030$} & 0.777{\scriptsize$\,\pm\,0.013$} & 0.880{\scriptsize$\,\pm\,0.011$} \\
\midrule
BBBP & AttentiveFP/A & 0.834{\scriptsize$\,\pm\,0.013$} & 0.658{\scriptsize$\,\pm\,0.020$} & 0.904{\scriptsize$\,\pm\,0.008$} & 0.918{\scriptsize$\,\pm\,0.005$} & 0.890{\scriptsize$\,\pm\,0.011$} & 0.291{\scriptsize$\,\pm\,0.040$} & 0.937{\scriptsize$\,\pm\,0.002$} & 0.711{\scriptsize$\,\pm\,0.013$} \\
 & AttentiveFP/F & 0.819{\scriptsize$\,\pm\,0.028$} & 0.591{\scriptsize$\,\pm\,0.019$} & 0.896{\scriptsize$\,\pm\,0.018$} & 0.897{\scriptsize$\,\pm\,0.005$} & 0.897{\scriptsize$\,\pm\,0.038$} & 0.186{\scriptsize$\,\pm\,0.042$} & 0.915{\scriptsize$\,\pm\,0.014$} & 0.676{\scriptsize$\,\pm\,0.036$} \\
 & GAT/A & 0.840{\scriptsize$\,\pm\,0.024$} & 0.557{\scriptsize$\,\pm\,0.003$} & 0.910{\scriptsize$\,\pm\,0.016$} & 0.891{\scriptsize$\,\pm\,0.001$} & 0.931{\scriptsize$\,\pm\,0.032$} & 0.140{\scriptsize$\,\pm\,0.020$} & 0.883{\scriptsize$\,\pm\,0.009$} & 0.510{\scriptsize$\,\pm\,0.029$} \\
 & GAT/F & 0.805{\scriptsize$\,\pm\,0.018$} & 0.595{\scriptsize$\,\pm\,0.041$} & 0.887{\scriptsize$\,\pm\,0.012$} & 0.898{\scriptsize$\,\pm\,0.012$} & 0.876{\scriptsize$\,\pm\,0.027$} & 0.176{\scriptsize$\,\pm\,0.065$} & 0.902{\scriptsize$\,\pm\,0.008$} & 0.625{\scriptsize$\,\pm\,0.032$} \\
 & GIN/A & 0.836{\scriptsize$\,\pm\,0.035$} & 0.569{\scriptsize$\,\pm\,0.034$} & 0.907{\scriptsize$\,\pm\,0.023$} & 0.895{\scriptsize$\,\pm\,0.009$} & 0.921{\scriptsize$\,\pm\,0.053$} & 0.163{\scriptsize$\,\pm\,0.050$} & 0.859{\scriptsize$\,\pm\,0.035$} & 0.504{\scriptsize$\,\pm\,0.086$} \\
 & GIN/F & 0.812{\scriptsize$\,\pm\,0.024$} & 0.595{\scriptsize$\,\pm\,0.032$} & 0.891{\scriptsize$\,\pm\,0.017$} & 0.898{\scriptsize$\,\pm\,0.009$} & 0.885{\scriptsize$\,\pm\,0.036$} & 0.183{\scriptsize$\,\pm\,0.051$} & 0.924{\scriptsize$\,\pm\,0.006$} & 0.676{\scriptsize$\,\pm\,0.017$} \\
 & PNA/A & 0.849{\scriptsize$\,\pm\,0.022$} & 0.569{\scriptsize$\,\pm\,0.012$} & 0.916{\scriptsize$\,\pm\,0.014$} & 0.894{\scriptsize$\,\pm\,0.003$} & 0.939{\scriptsize$\,\pm\,0.031$} & 0.175{\scriptsize$\,\pm\,0.027$} & 0.906{\scriptsize$\,\pm\,0.024$} & 0.621{\scriptsize$\,\pm\,0.063$} \\
 & PNA/F & 0.857{\scriptsize$\,\pm\,0.012$} & 0.547{\scriptsize$\,\pm\,0.019$} & 0.921{\scriptsize$\,\pm\,0.007$} & 0.885{\scriptsize$\,\pm\,0.005$} & 0.961{\scriptsize$\,\pm\,0.017$} & 0.151{\scriptsize$\,\pm\,0.057$} & 0.935{\scriptsize$\,\pm\,0.005$} & 0.715{\scriptsize$\,\pm\,0.031$} \\
\bottomrule
\end{tabular}}
\end{table}

\begin{table}[h]
  \centering
  \caption{Regression test metrics (A: atom, F: fragment). Abbreviations:
  r2: coefficient of determination; rmse: root-mean-square error and
  mae: mean absolute error, both in target units; rto\_r2 ($r_0^2$): $R^2$ of a 
  zero-intercept fit of observed on predicted values; ccc the concordance 
  correlation coefficient \citep{lin_concordance_1989}; roy\_c and roy\_c\_inv:
Roy's $r_m^2 = r^2(1-\sqrt{|r^2-r_0^2|})$ and its inverse $r_m'^2$
computed with observed and predicted values swapped
\citep{roy_on_2014}; delta ($\Delta r_0^2$): the Golbraikh--Tropsha
statistic $|r_0^2-r_0'^2|$ \citep{golbraikh_beware_2002}. Higher
is better except for rmse, mae and delta, with conventional
acceptance thresholds $r_m^2 > 0.5$ and $\Delta r_0^2 < 0.3$.}
  \label{tab:reg}
  \resizebox{\linewidth}{!}{\begin{tabular}{llcccccccc}
\toprule
Dataset & Model / Enc. & r2 & rmse & mae & rto\_r2 & ccc & roy\_c & roy\_c\_inv & delta \\
\midrule
\midrule
BACE (reg) & AttentiveFP/A & 0.686{\scriptsize$\,\pm\,0.039$} & 0.750{\scriptsize$\,\pm\,0.047$} & 0.579{\scriptsize$\,\pm\,0.034$} & 0.720{\scriptsize$\,\pm\,0.028$} & 0.846{\scriptsize$\,\pm\,0.013$} & 0.579{\scriptsize$\,\pm\,0.085$} & 0.601{\scriptsize$\,\pm\,0.062$} & 0.032{\scriptsize$\,\pm\,0.013$} \\
 & AttentiveFP/F & 0.730{\scriptsize$\,\pm\,0.016$} & 0.673{\scriptsize$\,\pm\,0.020$} & 0.499{\scriptsize$\,\pm\,0.017$} & 0.743{\scriptsize$\,\pm\,0.011$} & 0.861{\scriptsize$\,\pm\,0.007$} & 0.655{\scriptsize$\,\pm\,0.051$} & 0.659{\scriptsize$\,\pm\,0.040$} & 0.026{\scriptsize$\,\pm\,0.017$} \\
 & GAT/A & 0.620{\scriptsize$\,\pm\,0.026$} & 0.827{\scriptsize$\,\pm\,0.028$} & 0.611{\scriptsize$\,\pm\,0.028$} & 0.671{\scriptsize$\,\pm\,0.010$} & 0.818{\scriptsize$\,\pm\,0.009$} & 0.483{\scriptsize$\,\pm\,0.047$} & 0.547{\scriptsize$\,\pm\,0.041$} & 0.010{\scriptsize$\,\pm\,0.007$} \\
 & GAT/F & 0.606{\scriptsize$\,\pm\,0.074$} & 0.810{\scriptsize$\,\pm\,0.074$} & 0.619{\scriptsize$\,\pm\,0.052$} & 0.646{\scriptsize$\,\pm\,0.054$} & 0.814{\scriptsize$\,\pm\,0.031$} & 0.496{\scriptsize$\,\pm\,0.093$} & 0.571{\scriptsize$\,\pm\,0.077$} & 0.010{\scriptsize$\,\pm\,0.006$} \\
 & GIN/A & 0.551{\scriptsize$\,\pm\,0.029$} & 0.898{\scriptsize$\,\pm\,0.029$} & 0.687{\scriptsize$\,\pm\,0.024$} & 0.621{\scriptsize$\,\pm\,0.018$} & 0.797{\scriptsize$\,\pm\,0.009$} & 0.410{\scriptsize$\,\pm\,0.047$} & 0.515{\scriptsize$\,\pm\,0.033$} & 0.021{\scriptsize$\,\pm\,0.014$} \\
 & GIN/F & 0.558{\scriptsize$\,\pm\,0.169$} & 0.848{\scriptsize$\,\pm\,0.150$} & 0.640{\scriptsize$\,\pm\,0.095$} & 0.603{\scriptsize$\,\pm\,0.171$} & 0.800{\scriptsize$\,\pm\,0.056$} & 0.451{\scriptsize$\,\pm\,0.154$} & 0.552{\scriptsize$\,\pm\,0.076$} & 0.058{\scriptsize$\,\pm\,0.078$} \\
 & PNA/A & 0.715{\scriptsize$\,\pm\,0.013$} & 0.716{\scriptsize$\,\pm\,0.017$} & 0.551{\scriptsize$\,\pm\,0.017$} & 0.743{\scriptsize$\,\pm\,0.018$} & 0.858{\scriptsize$\,\pm\,0.007$} & 0.596{\scriptsize$\,\pm\,0.015$} & 0.631{\scriptsize$\,\pm\,0.024$} & 0.021{\scriptsize$\,\pm\,0.015$} \\
 & PNA/F & 0.680{\scriptsize$\,\pm\,0.026$} & 0.732{\scriptsize$\,\pm\,0.030$} & 0.560{\scriptsize$\,\pm\,0.028$} & 0.722{\scriptsize$\,\pm\,0.011$} & 0.845{\scriptsize$\,\pm\,0.010$} & 0.545{\scriptsize$\,\pm\,0.050$} & 0.597{\scriptsize$\,\pm\,0.043$} & 0.013{\scriptsize$\,\pm\,0.006$} \\
\midrule
PTR1 & AttentiveFP/A & -0.216{\scriptsize$\,\pm\,0.077$} & 1.179{\scriptsize$\,\pm\,0.038$} & 1.025{\scriptsize$\,\pm\,0.036$} & -0.072{\scriptsize$\,\pm\,0.057$} & 0.267{\scriptsize$\,\pm\,0.027$} & -0.132{\scriptsize$\,\pm\,0.044$} & -0.937{\scriptsize$\,\pm\,0.080$} & 1.077{\scriptsize$\,\pm\,0.125$} \\
 & AttentiveFP/F & -0.115{\scriptsize$\,\pm\,0.091$} & 1.112{\scriptsize$\,\pm\,0.046$} & 0.950{\scriptsize$\,\pm\,0.042$} & 0.128{\scriptsize$\,\pm\,0.053$} & 0.439{\scriptsize$\,\pm\,0.033$} & -0.054{\scriptsize$\,\pm\,0.046$} & -0.254{\scriptsize$\,\pm\,0.049$} & 0.398{\scriptsize$\,\pm\,0.039$} \\
 & GAT/A & -0.092{\scriptsize$\,\pm\,0.110$} & 1.117{\scriptsize$\,\pm\,0.057$} & 0.937{\scriptsize$\,\pm\,0.057$} & -0.065{\scriptsize$\,\pm\,0.094$} & 0.261{\scriptsize$\,\pm\,0.061$} & -0.072{\scriptsize$\,\pm\,0.090$} & -1.199{\scriptsize$\,\pm\,0.212$} & 1.282{\scriptsize$\,\pm\,0.145$} \\
 & GAT/F & -1.912{\scriptsize$\,\pm\,0.522$} & 1.792{\scriptsize$\,\pm\,0.159$} & 1.221{\scriptsize$\,\pm\,0.110$} & -1.826{\scriptsize$\,\pm\,0.535$} & 0.229{\scriptsize$\,\pm\,0.034$} & -1.414{\scriptsize$\,\pm\,0.498$} & -0.075{\scriptsize$\,\pm\,0.063$} & 1.788{\scriptsize$\,\pm\,0.561$} \\
 & GIN/A & -0.986{\scriptsize$\,\pm\,0.155$} & 1.507{\scriptsize$\,\pm\,0.058$} & 1.148{\scriptsize$\,\pm\,0.036$} & -0.746{\scriptsize$\,\pm\,0.189$} & 0.298{\scriptsize$\,\pm\,0.035$} & -0.513{\scriptsize$\,\pm\,0.129$} & -0.119{\scriptsize$\,\pm\,0.036$} & 0.631{\scriptsize$\,\pm\,0.169$} \\
 & GIN/F & -0.389{\scriptsize$\,\pm\,0.273$} & 1.236{\scriptsize$\,\pm\,0.124$} & 0.999{\scriptsize$\,\pm\,0.082$} & -0.282{\scriptsize$\,\pm\,0.299$} & 0.379{\scriptsize$\,\pm\,0.053$} & -0.294{\scriptsize$\,\pm\,0.229$} & -0.174{\scriptsize$\,\pm\,0.059$} & 0.273{\scriptsize$\,\pm\,0.254$} \\
 & PNA/A & -0.292{\scriptsize$\,\pm\,0.059$} & 1.216{\scriptsize$\,\pm\,0.028$} & 1.023{\scriptsize$\,\pm\,0.041$} & -0.150{\scriptsize$\,\pm\,0.043$} & 0.300{\scriptsize$\,\pm\,0.044$} & -0.181{\scriptsize$\,\pm\,0.026$} & -0.603{\scriptsize$\,\pm\,0.219$} & 0.565{\scriptsize$\,\pm\,0.338$} \\
 & PNA/F & -0.105{\scriptsize$\,\pm\,0.117$} & 1.107{\scriptsize$\,\pm\,0.059$} & 0.942{\scriptsize$\,\pm\,0.063$} & 0.133{\scriptsize$\,\pm\,0.031$} & 0.377{\scriptsize$\,\pm\,0.040$} & -0.041{\scriptsize$\,\pm\,0.058$} & -0.533{\scriptsize$\,\pm\,0.046$} & 0.916{\scriptsize$\,\pm\,0.180$} \\
\bottomrule
\end{tabular}
}
\end{table}

\section{Sensitivity of the aggregated baseline}
\label{app:conventions}
As mentioned in the main text, summing atom attributions into fragments 
requires two choices that no theory fixes: whether to aggregate signed 
attributions or their magnitudes, and whether to divide by the fragment's 
atom count.
Table~\ref{tab:conventions} gives the top-1 pharmacophore hit rate under
all four combinations, with Integrated Gradients on the representative
architecture per case study. The baseline spans 0.223--0.426 on BACE,
bracketing the native fragment model, and 0.696--0.783 on PTR1, where
the native model ties the upper end because three of the five values
coincide. The native model admits no such choice and reports one value.
\begin{table}[h]
  \centering
  \caption{Fragment hit rate of the aggregated atom-level baseline under
  each aggregation convention, against the convention-free native
  fragment model.}
  \label{tab:conventions}
  \begin{tabular}{lcc}
\toprule
Aggregation & BACE (cls) & PTR1 \\
\midrule
Signed, sum (SME default) & 0.399 & 0.783 \\
Signed, size-normalised & 0.223 & 0.696 \\
Absolute, sum & 0.426 & 0.783 \\
Absolute, size-normalised & 0.257 & 0.739 \\
\midrule
\emph{Native fragment model} & 0.277 & 0.783 \\
\bottomrule
\end{tabular}

\end{table}

\section{Matching pharmacophores to fragments}
\label{app:matching}
Reference labels are assigned by matching each SMARTS pattern on the
intact molecule and crediting a fragment with a class when every atom of
a match lies inside it (Section~\ref{sec:explain}). The obvious alternative, 
matching the pattern against the fragment's own SMILES, is not equivalent.
Breaking a bond leaves an open valence that RDKit fills with hydrogen,
so patterns constrained on hydrogen count, such as
\texttt{[NX3;H2,H1]}, \texttt{[OX2H]} and \texttt{[CX4;H2,H3]}, are
evaluated against an environment the atom does not have in the parent molecule.

The two protocols disagree on 18.5\% of the labels the isolated version assigns 
on the BACE test split: a substituted amine cut from its
molecule reads as a primary amine and matches \texttt{basic\_amine}
(19 of 74 such labels), and cleaved carbons read as terminal methyl groups and
match the S1 alkyl class (6 of 35). The error runs the other way too:
A lone nitrogen fragment reads as ammonia and fails
\texttt{[NX3;H2,H1]} even where the parent molecule carries a genuine
basic amine at that atom, which accounts for most of the 73 labels the
isolated protocol misses. PTR1 is unaffected: the isolated protocol
assigns 23 labels on its test split, the parent-molecule protocol
agrees with all of them, and it adds only two labels that the isolated
version misses. Its classes are ring systems, which a BRICS decomposition
typically leaves intact, so the hydrogen-count sensitivity that drives the 
BACE disagreement does not affect the results. We report the parent-molecule
protocol throughout.

\section{Scaffold-split and graph statistics}
\label{app:scaffold}
Table~\ref{tab:scaffold} reports split sizes, scaffold counts, BRICS
exclusion, and graph sizes.
\begin{table}[h]
  \centering
  \caption{Scaffold, BRICS-exclusion, and graph-size statistics.}
  \label{tab:scaffold}
  \resizebox{\linewidth}{!}{\begin{tabular}{lrrrrrrr}
\toprule
Dataset & Train & Val & Test & Scaf. & Singl. & BRICS excl. & Avg. atom/frag. nodes \\
\midrule
BACE (cls) & 1210 & 151 & 151 & 649 & 452 & 5 (0.3\%) & 34.1/7.2 \\
BACE (reg) & 1210 & 151 & 151 & 649 & 452 & 5 (0.3\%) & 34.1/7.2 \\
BBBP & 1572 & 196 & 197 & 1020 & 776 & 254 (12.9\%) & 24.0/4.5 \\
PTR1 & 200 & 25 & 25 & 119 & 82 & 11 (4.4\%) & 23.3/4.4 \\
\bottomrule
\end{tabular}
}
\end{table}
\newpage

\section{Sources of the PTR1 dataset}
\label{app:ptr1sources}
The 250 PTR1 compounds were curated by the authors from 16 primary
medicinal-chemistry publications listed in Table~\ref{tab:ptr1sources}. 
The manuscripts report 303 measurements between them, of which 53 are dropped
during curation because the activity is qualitative due to assay detection
limits rather than representing a fully qualified numeric $K_\text{i}$: 
a ``$>$'' bound only constrains the true $K_\text{i}$ from below, and readings marked 
inactive, insoluble or contaminated carry no usable value. Those 53 include 
the 16 entries that report no numeric $K_\text{i}$ at all. No compound is 
removed as a duplicate, and every source survives curation. Potencies are 
taken as $K_\text{i}$ rather than IC$_{50}$ because several compounds
were published directly reporting only $K_\text{i}$ values. For 155 compound 
activities reported originally as IC$_{50}$ values, the assay conditions
(enzyme concentration, substrate concentration, and enzyme K$_\text{m}$) 
were largely known and IC$_{50}$ values could thus be converted into $K_\text{i}$ 
values according to the Cheng-Prussoff equation (eq.~\ref{eq:ic50toki}),
assuming competitive inhibition.

\begin{equation}
  K_i = \frac{IC_{50}}{(1 + \frac{[S]}{K_m})}
 \label{eq:ic50toki}
\end{equation}

[S] corresponds to the substrate concentration during the IC$_{50}$
determination and K$_\text{m}$ to the Michaelis-Menten constant of the 
PTR1 enzyme in the absence of an inhibitor. As a result, the inhibition 
constant $K_\text{i}$ can be reported for 287 of the 303 entries.

\begin{table}[h]
  \centering
  \caption{Primary sources of the PTR1 dataset, ordered by the number
  of compounds each contributes after curation. ``Reported'' is the
  number of measurements in the source table and ``Curated'' the
  number surviving the removal of qualitative or unavailable readings.}
  \label{tab:ptr1sources}
  \begin{tabular}{llrr}
  \toprule
  Source & Year & Reported & Curated \\
  \midrule
  Khalaf et al. \citep{khalaf_structurebased_2014}      & 2014 &  71 &  46 \\
  Linciano et al. \citep{linciano_enhancement_2019}     & 2019 &  35 &  35 \\
  P\"ohner et al. \citep{pohner_multitarget_2022}       & 2022 &  34 &  34 \\
  Borsari et al. \citep{borsari_profiling_2016}         & 2016 &  42 &  29 \\
  Spinks et al. \citep{spinks_design_2011}              & 2011 &  26 &  22 \\
  Santucci et al. \citep{santucci_repurposing_2021}     & 2021 &  14 &  14 \\
  Tulloch et al. \citep{tulloch_structurebased_2010}    & 2010 &  18 &  13 \\
  Linciano et al. \citep{linciano_identification_2020}  & 2020 &  15 &  13 \\
  Linciano et al. \citep{linciano_exploiting_2017}      & 2017 &  11 &  10 \\
  Mpamhanga et al. \citep{mpamhanga_one_2009}           & 2009 &  11 &  10 \\
  Tassone et al. \citep{tassone_evidence_2021}          & 2021 &   7 &   7 \\
  Possart et al. \citep{possart_sesquiterpene_2022}     & 2022 &   5 &   5 \\
  Landi et al. \citep{landi_structural_2019}            & 2019 &   5 &   5 \\
  Di Pisa et al. \citep{dipisa_chroman_2017}            & 2017 &   4 &   3 \\
  Dawson et al. \citep{dawson_highresolution_2010}      & 2010 &   4 &   3 \\
  Landi et al. \citep{landi_highresolution_2020}        & 2020 &   1 &   1 \\
  \midrule
  Total                                                 &      & 303 & 250 \\
  \bottomrule
\end{tabular}

\end{table}

\section{PTR1 under data scarcity}
\label{app:scarcity}
PTR1 is the only included target with the data profile of an early-stage
program, the regime that motivates low-data and few-shot approaches to
molecular property prediction \citep{altaetran_low_2017}, and it
degrades the evaluation in ways worth separating from the
representation question. Model selection is the first major limitation:
validation-to-test CCC gaps run from 0.24 to 0.60, the largest of the
four tasks, because 150 search trials scored on 25 validation
molecules select on noise as much as on signal. Calibration fails too.
Every PTR1 test $R^2$ is negative, so no configuration beats a
constant predictor in squared error, while CCC stays positive by
rewarding rank-and-scale agreement. Absolute potency prediction is not
available at this sample size, and we do not claim it.

The explanations hold up better than the predictions. The attributions
remain pharmacophorically sensible even where the models are not
predictive, and they recover the pterin head at 3.2 times chance. A
model can therefore attribute to correct chemical substructures while
generalizing poorly, which is a caution about plausibility metrics in
general: they measure agreement with prior chemistry, not fidelity to
the model, and the two come apart where data are thin. The
explanation metrics also lose the resolution to compare approaches,
since all three select the same molecules, so plausibility evaluation
needs a larger test set than accuracy evaluation does. The
representation change itself costs nothing at this scale: the
five-fold node reduction holds (23.3 to 4.4), and the
fragment encoding wins on three of four architectures.

The PTR1 agreement reported in Section~\ref{sec:expq} should be 
interpreted with caution. The three approaches do not merely score alike, 
they select the same 18 of 23 molecules. With 4.4 fragments per molecule
and roughly a quarter of them pharmacophoric, the substrate-mimicking 
pterin head is likely to emerge as the top-ranked fragment under any of the
rankings, so the metric may saturate and offer limited resolution at
this scale. The agreement is therefore consistent with equivalence
among the approaches, but a larger or more diverse test set would be
needed to establish it.

The 0.783 ceiling is itself split-dependent: over all 239 decomposable
molecules it falls to 0.481, because scaffold splitting sends every
multi-member scaffold group to training and leaves a test set of
singletons. The above-chance recovery is nonetheless robust to how the
pharmacophore classes are drawn. Under a deliberately permissive
registry that counts any aromatic ring as a stacking element, chance
rises to 0.490 and the hit rate to 0.957, which still separates at
95\% confidence.

\section{Additional attribution maps}
\label{app:maps}
\begin{figure}[h]
  \centering
  \includegraphics[width=\linewidth]{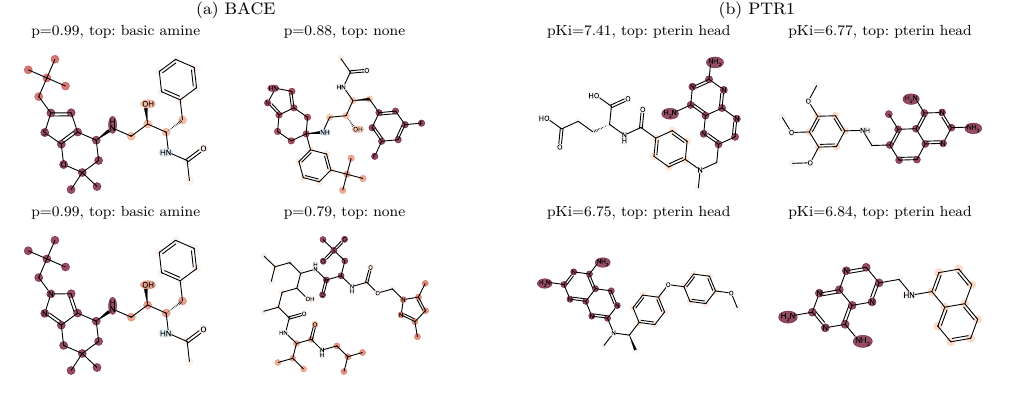}
  \caption{Native fragment-level IG attributions, each map scaled to
  its own maximum. (a) BACE actives; titles give the predicted
  probability and the top fragment's pharmacophore class. (b) PTR1
  test molecules whose top-ranked fragment is pharmacophoric; titles
  give the predicted pK$_\text{i}$ and the top fragment's
  pharmacophore class. Class labels mark fragments carrying a
  SAR-defined recognition element (Section~\ref{sec:explain}); for
  BACE, the hydroxyethylamine motif spans a BRICS cut, so no single
  fragment can carry it whole.}
  \label{fig:maps}
\end{figure}

Figure~\ref{fig:maps} shows native fragment-level IG attributions for
additional correctly-predicted BACE inhibitors in panel~(a), and the
same for PTR1 test molecules in panel~(b). The PTR1 maps make the
decomposition argument behind Figure~\ref{fig:hitrate} visible: the
pterin head survives BRICS as a single fragment and is the
top-attributed one in all four molecules. This is the behavior BACE
cannot exhibit, because its hydroxyethylamine motif spans a BRICS cut.

\end{document}